\documentclass[final]{cvpr}

\usepackage{times}
\usepackage{epsfig}
\usepackage{graphicx}
\usepackage{amsmath}
\usepackage{amssymb}

\usepackage[pagebackref=true,breaklinks=true,colorlinks,bookmarks=false]{hyperref}

\usepackage{booktabs}
\usepackage{multirow}
\usepackage{array}
\usepackage{arydshln}
\usepackage{diagbox}
\usepackage{bbding}
\usepackage{float}
\usepackage{stmaryrd}   
\usepackage{graphbox}
\usepackage{bm}

\begin{document}

\newcolumntype{I}{!{\vrule width 1.2pt}}
\newlength\savedwidth
\newcommand\whline{\noalign{\global\savedwidth\arrayrulewidth
                            \global\arrayrulewidth 1.1pt}
                   \hline
                   \noalign{\global\arrayrulewidth\savedwidth}}
\newlength\savewidth
\newcommand\shline{\noalign{\global\savewidth\arrayrulewidth
                            \global\arrayrulewidth 1.1pt}
                   \hline
                   \noalign{\global\arrayrulewidth\savewidth}}
\definecolor{darkergreen}{RGB}{21, 152, 56}
\newcommand\greenp[1]{\textcolor{darkergreen}{(#1)}}
\newcommand{\tabincell}[2]{\begin{tabular}{@{}#1@{}}#2\end{tabular}}  




\def\cvprPaperID{****} 
\def\confYear{CVPR 2021}

\title{DisCo: Remedy Self-supervised Learning on Lightweight Models with Distilled Contrastive Learning}

\author{Yuting Gao$^{1}\;$\footnotemark[1] \quad Jia-Xin Zhuang$^{1,2}$\thanks{The first two authors contributed equally. This work was done when Jia-Xin Zhuang was intern in Tencent Youtu Lab.} \quad Shaohui Lin$^3$  \quad Hao Cheng$^{1}$ \quad Xing Sun$^1$ \quad Ke Li \thanks{Corresponding Author, tristanli.sh@gmail.com} \quad Chunhua Shen$^{4}$\\ 
$^1$Tencent Youtu Lab \quad $^2$Hong Kong University of Science and Technology
 \quad \\ $^3$East China Normal University \quad $^4$Zhejiang University\\
{\tt\small \{yutinggao\}@tencent.com}\\ {\tt\small \{lincolnz9511,shaohuilin007,louischeng369,winfred.sun,tristanli,shchhshen\}@gmail.com}\\
}

\maketitle

\begin{abstract}

While Self-Supervised representation Learning (SSL) has received widespread attention from the community, recent researches argue that its performance will suffer a cliff fall when the model size decreases. Since current SSL methods mainly rely on contrastive learning to train the network, in this work, we propose a simple yet effective method termed \textbf{Dis}tilled \textbf{Co}ntrastive Learning (DisCo) to ease the issue. Specifically, we find the final inherent embedding of the mainstream SSL methods contains the most fruitful information, and propose to distill the final embedding to maximally transmit a teacher's knowledge to a lightweight model by constraining the last embedding of the student to be consistent with that of the teacher. In addition, we find that there exists a phenomenon termed Distilling BottleNeck and propose to enlarge the embedding dimension to alleviate this problem. Since the MLP only exists during the SSL phase, our method does not introduce any extra parameter to lightweight models during the downstream task deployment. Experimental results demonstrate that our method surpasses the state-of-the-art on all lightweight models by a large margin. Particularly, when ResNet-101/ResNet-50 is used respectively as a teacher to teach EfficientNet-B0, the linear result of EfficientNet-B0 on ImageNet is improved by 22.1\% and 19.7\% respectively, which is very close to ResNet-101/ResNet-50 with much fewer parameters. Code is available at \url{https://github. com/Yuting-Gao/DisCo-pytorch}.

\end{abstract}

\section{Introduction}

\begin{figure}[ht]
    \centering
    \vspace{-5pt}
    \includegraphics[width=1.0\linewidth]{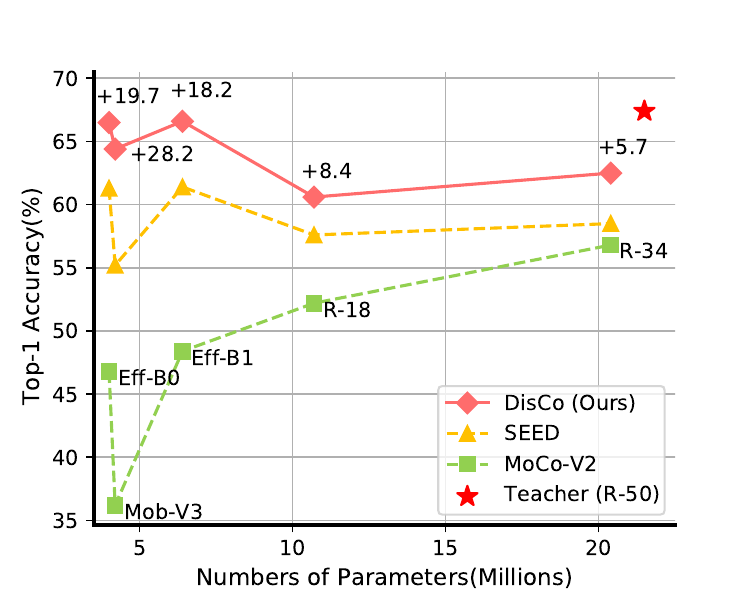}
    \vspace{-15pt}
    \caption{ImageNet top-1 linear evaluation accuracy on different network architectures. Our method significantly exceeds the result of using MoCo-V2  directly, and also surpasses the state-of-the-art SEED with a large margin. Particularly, the result of EfficientNet-B0 is quite close to the teacher ResNet-50, while the number of parameters of EfficientNet-B0 is only 16.3\% of ResNet-50. The improvement brought by DisCo is compared to MoCo-V2 baseline.}
    \vspace{-10pt}
    \label{fig:performance}
\end{figure}    

Deep learning has achieved great success in computer vision tasks, including image classification, object detection, and semantic segmentation. Such success relies heavily on manually labeled datasets, which are time-consuming and expensive to obtain. Therefore, more and more researchers begin to explore how to make better use of off-the-shelf unlabeled data. Among them, Self-supervised Learning (SSL) is an effective way to explore the information contained in the data itself by using proxy signals as supervision. Usually, after pre-training the network on massive unlabeled data with self-supervised methods and fine-tuning on downstream tasks, the performance of downstream tasks will be significantly improved. 
Hence, SSL has attracted widespread attention from the community, and many methods have been proposed \cite{komodakis2018unsupervised,noroozi2016unsupervised,doersch2015unsupervised,pathak2016context,chen2020simple,chen2020big, he2019moco,grill2020bootstrap}.
Among them, methods 
based on contrastive learning are becoming the mainstream due to their superior results. These methods are constantly refreshing the SOTA results with relatively large networks, but are unsatisfactory on some lightweight models at the same time. For example, the number of parameters of MobileNet-v3-Large/ResNet-152 is 5.2M/57.4M \cite{howard2019searching,he2016deep}, and the corresponding linear evaluation top-1 accuracy on ImageNet \cite{ILSVRC15} using MoCo-V2 \cite{chen2020mocov2} is 36.2\%/74.1\%. Compared to their fully supervised counterparts 75.2\%/78.57\%, the results of MobileNet-v3-Large is far from satisfying. Meanwhile in real scenarios, sometimes only lightweight models can be deployed due to the limited hardware resources. Therefore, improving the ability of self-supervised representation learning on small models is of great significance.

Knowledge distillation \cite{hinton2015distilling} is an effective way to transfer the knowledge learned by a large model (teacher) to a small model (student). Recently, some self-supervised representation learning methods use knowledge distillation to improve the efficacy of small models. SimCLR-V2 \cite{chen2020big} uses logits in the fine-tuning stage to transfer the knowledge in a task-specific way. CompRess \cite{koohpayegani2020compress} and SEED \cite{fang2021seed} mimic the similarity score distribution between a teacher and a student model over a dynamically maintained queue. Though distillation is effective, two factors affect the result prominently, \emph{i.e.}, \textit{which} knowledge is most needed by the student and \textit{how} to deliver it. In this work, we propose new insights towards these two aspects.

In the current mainstream contrastive learning based SSL methods, a multi-layer perceptron (MLP) is added after the encoder to obtain a low-dimensional embedding. Training loss and the accuracy evaluation are both performed on this embedding. We thus hypothesize that this final embedding contains the most fruitful knowledge and should be regarded as the first choice for knowledge transfer. To achieve this, we propose a simple yet effective \textbf{Dis}tilled \textbf{Co}ntrastive Learning (DisCo) framework to transfer this knowledge from large models to lightweight models in the pre-training stage. Specifically, 
DisCo takes the MLP embedding obtained by the teacher as the knowledge and injects it into the student by constraining the corresponding embedding of the student to be consistent with that of the teacher using MSE loss. 
In addition, we find that a budgeted dimension of the hidden layer in the MLP of the student may cause a knowledge transmission bottleneck. We term this phenomenon as \textit{Distilling Bottleneck} and present to enlarge the embedding dimension to alleviate this problem. This simple yet effective operation relates to the capability of model generalization in the setting of self-supervised learning from the Information BottleNeck \cite{tishby2000information} perspective. It is worth noting that our method only introduces a small number of additional parameters in the pre-training phase, but during the fine-tuning and deployment stage, there is no extra computational burden since the MLP layer is removed.

Experimental results demonstrate that DisCo can effectively transfer the knowledge from the teacher to the student, making the representations extracted by the student more generalized. Our approach is simple and incorporate it into existing contrastive based SSL methods can bring significant gains. Our contributions are summarized as follows: 

\begin{itemize}

\item We propose a simple yet effective self-supervised distillation method to boost the representation abilities of lightweight models.

\item We discover that there exists a phenomenon termed Distilling BottleNeck in the self-supervised distillation stage and propose to enlarge the embedding dimension to alleviate this problem.

\item We achieve state-of-the-art SSL results on lightweight models. Particularly, the linear evaluation results of EfficientNet-B0 \cite{tan2019efficientnet} on ImageNet is quite close to ResNet-101/ResNet-50, while the number of parameters of EfficientNet-B0 is only 9.4\%/16.3\% of ResNet-101/ResNet-50.

\end{itemize}

\section{Related Work}

\subsection{Self-supervised Learning}

Self-supervised learning (SSL) is a generic framework that learns high semantic patterns from the data itself without any tags from human beings. Current methods mainly rely on three paradigms, \textit{i.e.},  pretext tasks, contrastive based and clustering based.

\noindent \textbf{Pretext tasks.} 
Approaches based on pretext paradigm focus on designing more effective surrogate tasks, including Exampler-CNN \cite{dosovitskiy2015discriminative} that identifies whether patches are cropped from the same image, Rotation \cite{komodakis2018unsupervised} that predicts the rotation degree of the input image, Jigsaw \cite{noroozi2016unsupervised} that places the shuffled patches back to the original position, and Context encoder \cite{pathak2016context} that recovers the missing part of the input image conditioned on its surrounding.

\noindent \textbf{Contrastive based.} Contrastive based approaches have shown impressive performance on self-supervised representation learning, which enforce different views of the same input to be closer in feature space \cite{chen2020exploring,chen2020big,chen2020simple,henaff2020data,he2019moco,chen2020mocov2,grill2020bootstrap,wang2020enhancing,wang2020removing,zbontar2021barlow}. SimCLR \cite{chen2020simple,chen2020big} indicates that self-supervised learning can be boosted by applying strong data augmentation, training with larger batch size of negative samples, and adding projection head (MLP) after the global average pooling. However, SimCLR relies on very large batch size to achieve comparable performance, and cannot be applied widely to many real-world scenarios. MoCo \cite{he2019moco,chen2020mocov2} considers contrastive learning as a look-up dictionary, using a memory bank to maintain consistent representations of negative samples. Thus, MoCo can achieve superior performance without large batch size, which is more feasible to implement. BYOL \cite{grill2020bootstrap} introduces a predictor to one branch of the network to break the symmetry and avoid the trivial solution. DINO \cite{caron2021emerging} applies 
contrastive learning to vision transformers.

\noindent \textbf{Clustering based.} Clustering is one of the most promising approaches for unsupervised representation learning. DeepCluster \cite{caron2018deep} uses k-means assignments to generate pseudo-labels to iteratively group the features and update the weight of the network. DeeperCluster \cite{caron2019unsupervised} scales to large uncurated datasets to capture complementary statistics. 
Different from previous works, to maximize the mutual information between pseudo labels and input data, SeLa \cite{asano2020self} cast the pseudo-label assignment as an instance of optimal transport. SwAV \cite{caron2020unsupervised} formulates to map representations to prototype vectors, which is assigned online and is capable to scale to larger datasets.

Although the mainstream methods SimCLR-V2, MoCo-V2, BYOL and SwAV belong to different self-supervised categories, they have four things in common: 1) two views for one image, 2) two encoders for feature extraction, 3) two projection heads to map the representations into a lower dimension space, and 4) the two low-dimensional embeddings are regarded to be a pair of positive samples, which can be considered as a contrast process. However, all of these methods suffer a performance cliff fall on lightweight models, which is what we try to remedy in this work.


\subsection{Knowledge Distillation}
Knowledge distillation (KD) tries to transfer the knowledge from a larger teacher model to a smaller student model. 
According to the form of knowledge, it can be classified into three categories, logits-based, feature-based, and relation-based.

\noindent \textbf{Logits-based.} Logits refers to the output of the network classifier. KD \cite{hinton2015distilling} proposes to make the student mimic the logits of the teacher by minimizing the KL-divergence of the class distribution. 

\noindent \textbf{Feature-based.} Feature-based methods directly transfer the knowledge from the intermediate layers of the teacher to student. FitNets \cite{romero2014fitnets} regards the intermediate representations learned by the teacher as hints and transfers the knowledge to a thinner and deeper student through minimizing the mean square error between the representations. AT \cite{zagoruyko2016paying} proposes to use the spatial attention of the teacher as the knowledge and let the student pay attention to the area that the teacher is concerned about. SemCKD \cite{chen2020cross} adaptively selects the more appropriate representation pairs of the teacher and student. 

\noindent \textbf{Relation-based.} Relation-based approaches explore the relationship between data instead of the output of a single instance. RKD \cite{park2019relational} transfers the mutual relationship of the input data within one batch with distance-wise and angle-wise distillation loss from the teacher to the student. IRG \cite{liu2019knowledge} proposes to use the relationship the graph to further express the relational knowledge.

 
\subsection{SSL meets KD}

Recently, some works combine self-supervised learning and knowledge distillation.
CRD \cite{tian2019contrastive} introduces
a contrastive loss to transfer pair-wise relationship across different modalities. SSKD \cite{xu2020knowledge} lets the student mimic transformed data and self-supervision tasks to transfer richer knowledge. The above-mentioned works take self-supervision as an auxiliary task to further boost the process of knowledge distillation under fully supervised setting. CompRess\cite{koohpayegani2020compress} and SEED \cite{fang2021seed} tried to employ knowledge distillation as a means to improve the self-supervised visual representation learning capability of small models, which utilize the negative sample queue in MoCo \cite{he2019moco} to constrain the distribution of positive sample over negative samples of the student to be consistent with that of the teacher. However, CompRess and SEED heavily rely on MoCo framework, which means that a memory bank always has to be preserved during the distillation process. Our method also aims to boost the self-supervised representation learning ability on lightweight models by distilling, however, we do not restrict the self-supervised framework and are thus more flexible. Furthermore, our method surpass SEED with a large margin on all lightweight models under the same setting.


\section{Method}
In this section, we introduce the proposed \textit{Distilled Contrastive Learning} (DisCo) framework on lightweight models. We first give some preliminaries on contrastive based SSL and then introduce the overall architecture of DisCo and how DisCo transfers the knowledge from the teacher to the student. Finally, we present how DisCo can be combined with the existing contrastive based SSL methods.

\subsection{Preliminary on Contrastive Learning Based SSL}
Mainstream contrastive learning based SSL methods have four common characteristics.

\textbf{Two views:} one input image $x$ is transformed into two views ${v}$ and $v^{'}$ by two drastic data augmentation operations.

\textbf{Two encoders:} two augmented views are input to two encoders of the same structure, one is a learnable base encoder $s(\cdot)$ and the other $m(\cdot)$ is updated according to the base encoder, either shared or momentum updated. The encoder here can use any network architecture, such as the commonly used ResNet. Given an input image, the extracted representation obtained from the last global average pooling of the encoder is denoted as $Z$, and its dimension is $D$.

\begin{figure}[htbp]
    \centering
    \vspace{-5pt}
    {\includegraphics[align=c,width=1\linewidth]{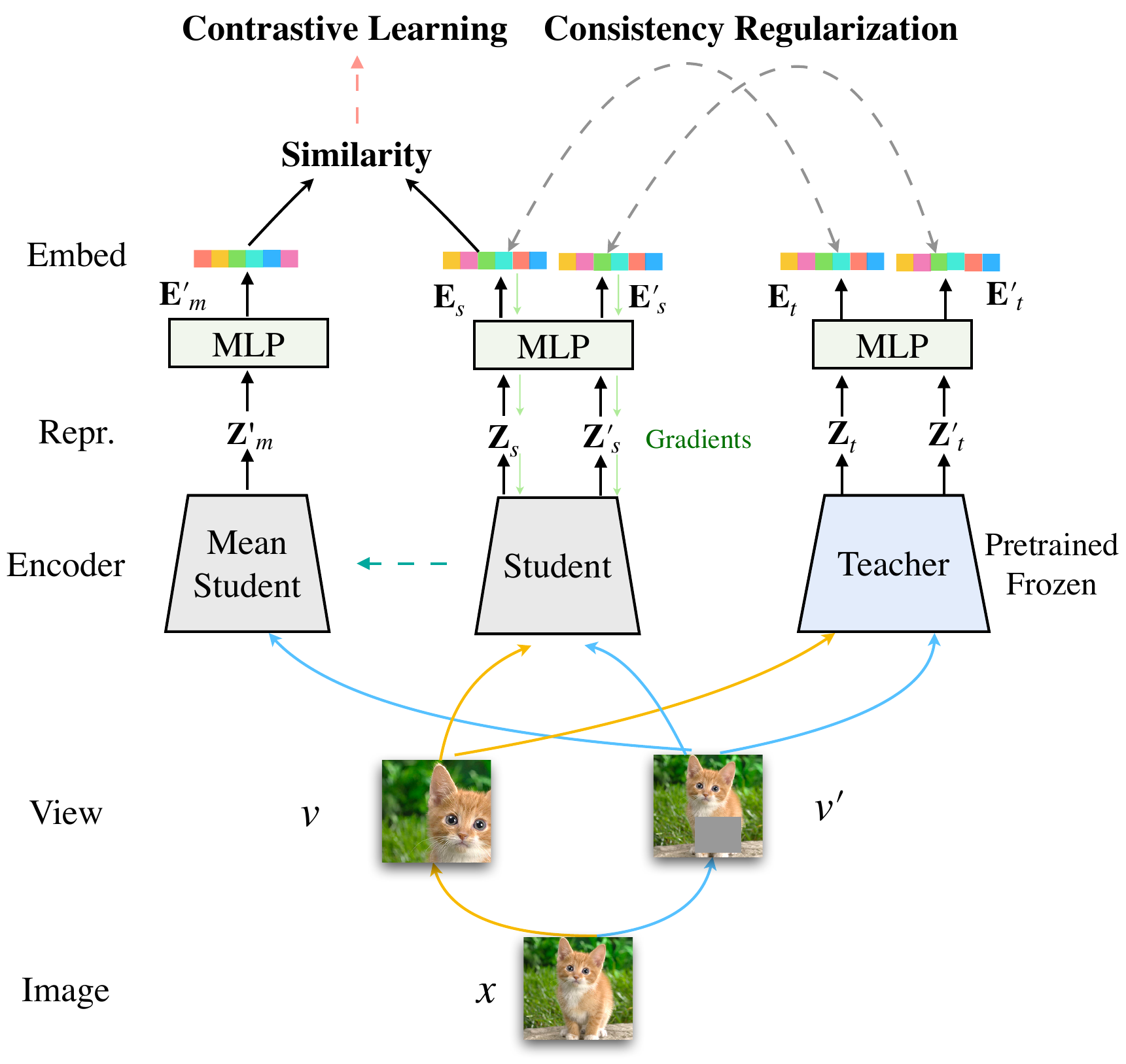}}
    \caption{
    The framework of the proposed method DisCo. One image is first transformed into two views by two drastic data augmentation operations. In addition to the original constrastive SSL part, a self-supervised pre-trained teacher is introduced, and the final embeddings obtained by the learnable student and the frozen teacher are required to be consistent for each view. Repr. stands for representation.}
    \vspace{-10pt}
    \label{fig:model}
\end{figure}

\textbf{Projection head:} both encoders are followed by a small projection head $p(\cdot)$ that maps the representation $Z$ to a low-dimensional embedding $E$, which contains several linear layers. This procedure can be formulated as $E = p(Z) =  W_{(n)}\cdots (\sigma(W _{(1)} Z))$, where $W$ is the weight parameter of the linear layer, $n$ is the number of layers, which is greater than or equal to 1, and $\sigma$ is the non-linear function ReLU. The importance the of projection head has been addressed in SimCLR-V2 and MoCo-V2. Following MoCo-V2, the default configuration of the projection head is two linear layers, in which the first layer maintains the original feature dimension D, and the second layer reduces the dimension to 128.


\textbf{Loss function:} after obtaining the final embeddings of these two views, they are regarded as a pair of positive samples to calculate the loss.

\subsection{Overall Architecture}

The framework of DisCo is shown in Figure \ref{fig:model}, consisting of three encoders followed by the projection head. The \textit{Student} $s(\cdot)$ in center is the encoder that we want to improve, the \textit{Mean Student} $m(\cdot)$ is updated according to $s(\cdot)$, and \textit{Teacher} $t(\cdot)$ is the self-supervised pre-trained large encoder that is used as teacher in distillation.


For each input image $x$, it is first transformed into two views $v$ and $v{'}$ by two drastic data augmentation operations. On the one hand, $v$ is input to $s(\cdot)$ and $t(\cdot)$, generating two representations $Z_s = s(v)$, $Z_t = t(v)$, then after the projection head, these two representations are mapped to low-dimensional embeddings, $E_s = p_s(Z_s)$, $E_t = p_t(Z_t)$ respectively. On the other hand, $v{'}$ is input to $s(\cdot)$, $m(\cdot)$ and $t(\cdot)$ simultaneously, after encoding and projecting, three low-dimensional vectors $E^{'}_s = p_s(s(v{'}))$, $E^{'}_m = p_m(m(v{'}))$, and $E^{'}_t = p_t(t(v{'}))$ are obtained.

$E^{'}_m$ and $E_s$ are the embeddings of two different views, which are regarded as a pair of positive samples and are pulled together in the existing SSL methods. $E_s$ and $E_t$, $E^{'}_s$ and $E^{'}_t$ are two pairs of embeddings of the student and the teacher of the same view, and each pair is constrained to be consistent during the distilling procedure.

\subsection{Distilling Procedure}

In most contrastive based SSL methods, the calculation of loss function and the evaluation of accuracy are performed at the final embedding vector $E$. Therefore, we hypothesize that the last embedding $E$ contains the most fruitful knowledge and should be primarily considered when distilling.

For a self-supervised pre-trained teacher model, we distill the knowledge in the last embedding into the student, that is, for view $v$ and view $v{'}$, the embedding vector output by the frozen teacher and the learnable student should be consistent. Specifically, we use a consistency regularization term to pull the embedding vector $E_s$ closer to $E_t$ and $E^{'}_s$ closer to $E^{'}_t$. Formally, 

\vspace{-10pt}

\begin{equation}
    \mathcal{L}_{dis} = {|| E_s - E_t||}^2 + {|| E^{'}_s - E^{'}_t||}^2 
\end{equation}

To verify that the embedding $E$ contains the most meaningful knowledge, we experiment with several other commonly used distillation schemes in Table \ref{tab:other-distillation}. The results prove that the knowledge we transmitted and the way it is transferred are indeed the most effective.

\begin{figure}[h]
    \centering
    \vspace{-10pt}
    \includegraphics[width=0.8\linewidth]{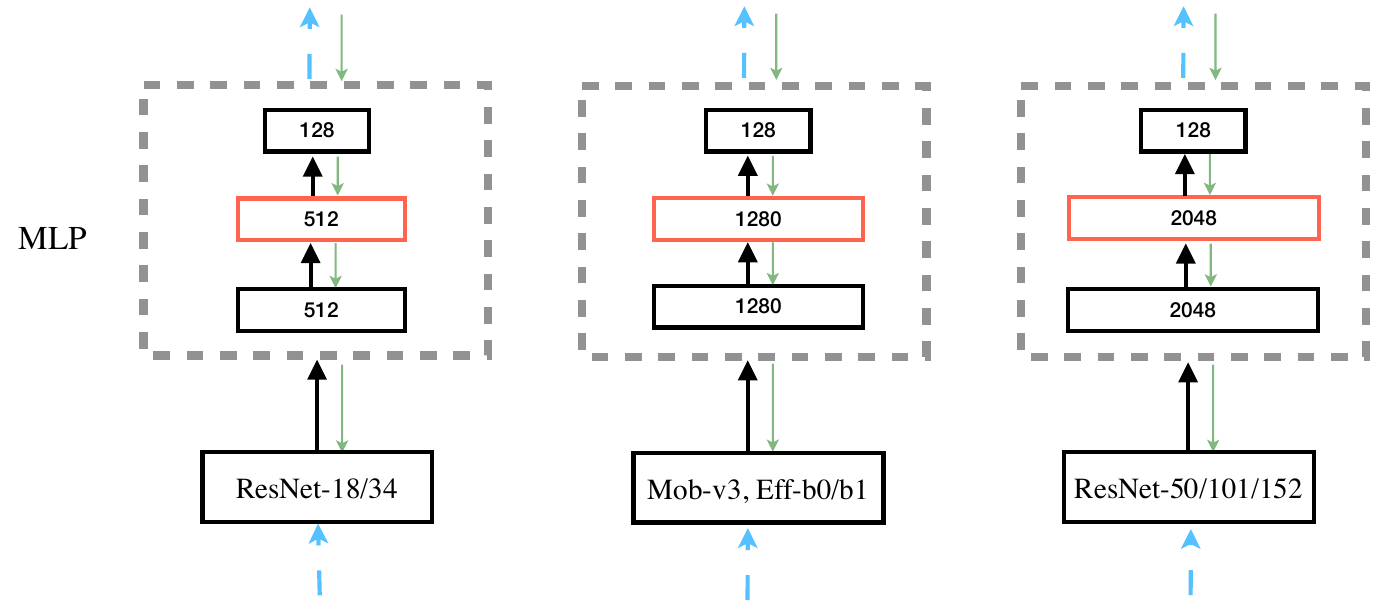}
    \vspace{-6pt}
    \caption{Default MLP of multiple networks.}
    \vspace{-10pt}
    \label{fig:mlp}
\end{figure}

\noindent\textbf{Distilling Bottleneck.} In our distillation experiment, we found an interesting phenomenon. When the encoder of the student is ResNet-18/34 and the default MLP configuration is adopted, that is, the dimension of embedding output by the encoder is projected from $D$ to $D$ and then to $128$, the results of DisCo is not satisfactory. We assume that this degradation is caused by the fact that the dimension of the hidden layer in the MLP is too small, and term this phenomenon as \textit{Distilling Bottleneck}. In Figure \ref{fig:mlp}, we exhibit the default configuration of the projection head of ResNet-18/34, EfficientNet-B0/B1, MobileNet-v3-Large,  and ResNet-50/101/152. It can be seen that the dimension of the hidden layer of ResNet-18/34 is too small compared to other networks.

To alleviate the Distilling Bottleneck problem, we expand the dimension of the hidden layer in MLP. It's worth noting that this operation only introduces a small number of parameters at the self-supervised distillation stage, and the MLP will be directly discarded during fine-tuning and deployment, which means no extra computational burden is brought. We experimentally verified that such a simple operation can bring significant gains in Table ~\ref{tab:AblationStudy}.

This operation can be explained from the Information Bottleneck (IB) \cite{tishby2000information} perspective. IB is utilized in \cite{shwartz2017opening,cheng2018evaluating} to understand how deep networks work by visualizing mutual information ($I(X;T)$ and $I(T;Y)$) in the information plane, where $I(X;T)$ is the mutual information between input and output, and $I(T;Y)$ is the mutual information between output and label. The training of deep networks can be described by two-phases: the first \textit{fitting phase}, where the network memorizes the information of input, resulting in the growth of $I(X;T)$ and $I(T;Y)$; the subsequent \textit{compression phase}, where the network removes irrelevant information of input for better generalization, resulting in the decrease of $I(X;T)$. Generally, in the \textit{compression phase}, $I(X;T)$ can present the model's capability of generalization while $I(T;Y)$ can present the model's capability of fitting label \cite{cheng2018evaluating}. We visualize the \textit{compression phase} of our model with different dimensions of the hidden layer in the pre-training distillation stage in the information plane on one downstream transferring classification task. The results in Figure \ref{fig:mlp-ib} shows two interesting phenomenons:

\emph{i}. Models with different dimensions of the hidden layer have very similar $I(T;Y)$, suggesting that models have the nearly equal capability of fitting the labels.

\emph{ii}. The Model with larger dimension in the hidden layer has smaller $I(X;T)$, suggesting a stronger capability of generalization.


These phenomenons show that MLP indeed relates to the capability of model generalization in the setting of self-supervised transfer learning. 




\subsection{Overall Objective Function}

The overall objective function is defined as follows:
\begin{equation}
    \label{eq:objective funciton}
    \mathcal{L} = \mathcal{L}_{dis} + \lambda \mathcal{L}_{co}
\end{equation}

where $\mathcal{L}_{dis}$ comes from the distillation part, $\mathcal{L}_{co}$ can be the contrastive loss of any SSL method, and $\lambda$ is a hyper-parameter that controls the weights of the distillation loss and contrastive loss. In our experiments, $\lambda$ is set to 1. Due to the simplicity of implementation, we use MoCo-V2 as the testbed in the experiments without additional explanation.

\section{Experiments}

\subsection{Settings}

\noindent \textbf{Dataset.} All the self-supervised pre-training experiments are conducted on ImageNet \cite{ILSVRC15}. 
For downstream classification tasks, experiments are carried out on Cifar10 and Cifar100 \cite{krizhevsky2009learning}. For downstream detection tasks, experiments are conducted on PASCAL VOC \cite{everingham2010pascal} and MS-COCO \cite{lin2014microsoft}, with train+val/test and train2017/val2017 for training/testing respectively. For downstream segmentation tasks, the proposed method is verified on MS-COCO.


\noindent \textbf{Teacher Encoders.} Four large encoders are used as teachers, ResNet-50(22.4M), ResNet-101(40.5M), ResNet-152(55.4M), and ResNet-50*2(55.5M), where X(Y) denotes that the encoder X has Y millions of parameters and the Y does not consider the linear layer.



\noindent \textbf{Student Encoders.} Five widely used small yet effective networks are used as student, EfficientNet-B0(4.0M), MobileNet-v3-Large(4.2M), EfficientNet-B1(6.4M), ResNet-18(10.7M) and ResNet-34(20.4M).


\noindent\textbf{Teacher Pre-training Setting.} ResNet-50/101/152 are pre-trained using MoCo-V2 with default hyper-parameters. Following SEED, ResNet-50 and ResNet-101 are trained for 200 epochs, and ResNet-152 is trained for 400 epochs. ResNet-50*2 is pre-trained by SwAV, which is an open-source model~\footnote{\url{https://github.com/facebookresearch/swav}} and trained for 800 epochs. 


\noindent\textbf{Self-supervised Distillation Setting.} The projection head of all the student networks has two linear layers, with the dimension being 2048 and 128. The configuration of the learning rate and optimizer is set the same as MoCo-V2, and without a specific statement, the model is trained for 200 epochs. During the distillation stage, the teacher is frozen.


\noindent \textbf{Student Fine-tuning Setting.} 
For linear evaluation on ImageNet, the student is fine-tuned for 100 epochs. Initial learning rate is 3 for EfficientNet-B0/EfficientNet-B1/MobileNet-v3-Large, and 30 for ResNet-18/34. For linear evaluation on Cifar10 and Cifar100, the initial learning rate is 3 and all the models are fine-tuned for 100 epochs. SGD is adopted as the optimizer and the learning rate is decreased by 10 at 60 and 80 epochs for linear evaluation. For downstream detection and segmentation tasks, following SEED \cite{fang2021seed}, all parameters are fine-tuned. For the detection task on VOC, the initial learning rate is 0.1 with 200 warm-up iterations and decays by 10 at 18k, 22.2k steps. The detector is trained for 48k steps with a batch size of 32. Following SEED, the scales of images are randomly sampled from [400, 800] during the training and is 800 at the inference. For the detection and instance segmentation on COCO, the model is trained for 180k iterations with the initial learning rate 0.11, and the scales of images are randomly sampled from [600, 800] during the training. 

\begin{table*}[ht]
    \centering
    \caption{ImageNet test accuracy (\%) using linear classification on different student architectures. $\Diamond$ denotes the teacher/student models are pre-trained with MoCo-V2, which is our implementation and \dag means the teacher is pre-trained by SwAV, which is an open-source model. When using R50*2 as the teacher, SEED distills 800 epochs while DisCo distills 200 epochs. Subscript in green represents the improvement compared to MoCo-V2 baseline. }
    \setlength{\tabcolsep}{1mm}{
    \begin{tabular}{c|c|cc|cc|cc|cc|cc}
        \whline
        \multirow{2}{*}{\textbf{Method}} & \multirow{2}{*}{\diagbox{\textbf{T}}{\textbf{S}}} & \multicolumn{2}{c|}{\textbf{Eff-b0}} & \multicolumn{2}{c|}{\textbf{Eff-b1}} & \multicolumn{2}{c|}{\textbf{Mob-v3}} & \multicolumn{2}{c|}{\textbf{R-18}} & \multicolumn{2}{c}{\textbf{R-34}} \\ 
        \cline{3-12}
        & & \textbf{T-1} & \textbf{T-5} & \textbf{T-1} & \textbf{T-5} & \textbf{T-1} & \textbf{T-5} & \textbf{T-1} & \textbf{T-5} & \textbf{T-1} & \textbf{T-5}  \\
        \hline
        \multicolumn{2}{c|}{\textbf{Supervised}} & 77.1 & 93.3 & 79.2 & 94.4 & 75.2 & -  & 72.1 & - & 75.0 & - \\
        \hline
        \multicolumn{2}{c|}{\textit{\textbf{Self-supervised}}} & & & & & & & & & & \\
        \multicolumn{2}{c|}{MoCo-V2 (Baseline)$\Diamond$} & 46.8 & 72.2 & 48.4 & 73.8 & 36.2 & 62.1 & 52.2 & 77.6 & 56.8 & 81.4\\
        \hline
        \textit{\textbf{SSL Distillation}} & & & & & & & & & & & \\
         SEED\cite{fang2021seed} & R-50 (67.4) & 61.3 & 82.7 & 61.4 & 83.1 & 55.2 & 80.3 & 57.6 & 81.8 & 58.5 & 82.6 \\
         \tabincell{c}{\textbf{DisCo (ours)} \\{}} & \tabincell{c}{R-50 (67.4)$\Diamond$\\ {}} &  \tabincell{c}{\textbf{66.5}\\ \greenp{19.7$\uparrow$}} &  \tabincell{c}{\textbf{87.6}\\ \greenp{15.4$\uparrow$}} &  \tabincell{c}{\textbf{66.6}\\ \greenp{18.2$\uparrow$}} &  \tabincell{c}{\textbf{87.5}\\ \greenp{13.7$\uparrow$}} &  \tabincell{c}{\textbf{64.4}\\ \greenp{28.2$\uparrow$}} &  \tabincell{c}{\textbf{86.2}\\ \greenp{24.1$\uparrow$}} &  \tabincell{c}{\textbf{60.6}\\ \greenp{8.4$\uparrow$}} &  \tabincell{c}{\textbf{83.7}\\ \greenp{6.1$\uparrow$}} &  \tabincell{c}{\textbf{62.5}\\ \greenp{5.7$\uparrow$}} &  \tabincell{c}{\textbf{85.4}\\ \greenp{4.0$\uparrow$}} \\
        \cdashline{1-12}
         SEED~\cite{fang2021seed} & R-101 (70.3) & 63.0 & 83.8 & 63.4 & 84.6 & 59.9 & 83.5 & 58.9 & 82.5 & 61.6 & 84.9 \\
         \tabincell{c}{\textbf{DisCo (ours)} \\ {}} & \tabincell{c}{R-101 (69.1)$\Diamond$\\ {}} &  \tabincell{c}{\textbf{68.9}\\ \greenp{22.1$\uparrow$}} & \tabincell{c}{\textbf{88.9}\\ \greenp{16.7$\uparrow$}} &  \tabincell{c}{\textbf{69.0}\\ \greenp{20.6$\uparrow$}} &  \tabincell{c}{\textbf{89.1}\\ \greenp{15.3$\uparrow$}} &  \tabincell{c}{\textbf{65.7}\\ \greenp{29.5$\uparrow$}} &  \tabincell{c}{\textbf{86.7}\\ \greenp{24.6$\uparrow$}} &  \tabincell{c}{\textbf{62.3}\\ \greenp{10.1$\uparrow$}} &  \tabincell{c}{\textbf{85.1}\\ \greenp{7.5$\uparrow$}} &  \tabincell{c}{\textbf{64.4}\\ \greenp{7.6$\uparrow$}} &  \tabincell{c}{\textbf{86.5}\\ \greenp{5.1$\uparrow$}}\\
         \cdashline{1-12}
         SEED~\cite{fang2021seed} & R-152 (74.2) & 65.3 & 86.0 & 67.3 & 86.9 & 61.4 & 84.6 & 59.5 & 83.3 & 62.7 & 85.8\\
         \tabincell{c}{\textbf{DisCo (ours)} \\{}} & \tabincell{c}{R-152 (74.1)$\Diamond$\\ {}} &  \tabincell{c}{\textbf{67.8}\\ \greenp{21.0$\uparrow$}} &  \tabincell{c}{\textbf{87.0}\\ \greenp{14.8$\uparrow$}} &  \tabincell{c}{\textbf{73.1}\\ \greenp{24.7$\uparrow$}} &  \tabincell{c}{\textbf{91.2}\\ \greenp{17.4$\uparrow$}} &  \tabincell{c}{\textbf{63.7}\\ \greenp{27.5$\uparrow$}} &  \tabincell{c}{\textbf{84.9}\\ \greenp{22.8$\uparrow$}} &  \tabincell{c}{\textbf{65.5}\\ \greenp{13.3$\uparrow$}} &  \tabincell{c}{\textbf{86.7}\\ \greenp{9.1$\uparrow$}} &  \tabincell{c}{\textbf{68.1}\\ \greenp{11.3$\uparrow$}} &  \tabincell{c}{\textbf{88.6}\\ \greenp{7.2$\uparrow$}}\\
         \cdashline{1-12}
         SEED~\cite{fang2021seed} & R50*2 (77.3\dag) & 67.6 & 87.4 & 68.0 & 87.6 & 68.2 & 88.2 & 63.0 & 84.9 & 65.7 & 86.8 \\
        \tabincell{c}{\textbf{DisCo (ours)} \\{}} & \tabincell{c}{R50*2 (77.3)\dag\\ {}} & \tabincell{c}{\textbf{69.1}\\ \greenp{22.3$\uparrow$}} & \tabincell{c}{\textbf{88.9}\\ \greenp{17.7$\uparrow$}} & \tabincell{c}{64.0\\ \greenp{15.6$\uparrow$}} & \tabincell{c}{84.6\\ \greenp{10.8$\uparrow$}} & \tabincell{c}{58.9\\ \greenp{22.7$\uparrow$}} & \tabincell{c}{81.4 \\ \greenp{19.3$\uparrow$}} & \tabincell{c}{\textbf{65.2} \\ \greenp{13$\uparrow$}} & \tabincell{c}{\textbf{86.8}\\ \greenp{9.2$\uparrow$}} &  \tabincell{c}{\textbf{67.6}\\ \greenp{10.8$\uparrow$}}&  \tabincell{c}{\textbf{88.6}\\ \greenp{7.2$\uparrow$}}\\
        \whline
    \end{tabular}
    }
    \label{tab:ImageNet-1k-baseline-our}
\end{table*}

\subsection{Linear Evaluation}
We conduct linear evaluation on ImageNet to validate the effectiveness of our method. As shown in Table \ref{tab:ImageNet-1k-baseline-our}, student models distilled by DisCo outperform the counterparts pre-trained by MoCo-V2 (Baseline) with a large margin. Besides, DisCo surpasses the state-of-the-art SEED over various student models with teacher ResNet-50/101/152 under the same setting, especially on MobileNet-v3-Large distilled by ResNet-50 with a difference of 9.2\% at top-1 accuracy. When using R50*2 as the teacher, SEED distills 800 epochs while DisCo still distills 200 epochs, but the results of EfficientNet-B0, ResNet-18 and, ResNet-34 using DisCo also exceed that of SEED. The performance on EfficientNet-B1 and MobileNet-v3-Large is closely related to the epochs of distillation. For example, when EfficientNet-B1 is distilled for 290 epochs, the top-1 accuracy becomes 70.4\%, which surpasses SEED and when MobileNet-v3-Large is distilled for 340 epochs, the top-1 accuracy becomes 64\%. We believe that when DisCo distills 800 epochs, the results will be further improved. Moreover, since CompRess uses a better teacher which trained 600 epochs longer and distills 400 epochs longer than SEED and ours, it's not fair to compare thus we do not report the result in the table. 
In addition, when DisCo uses a larger model as the teacher, the student will be further improved. For instance, using ResNet-152 instead of ResNet-50 as the teacher, ResNet-34 is improved from 62.5\% to 68.1\%. It's worth noting, when using ResNet-101/ResNet-50 as the teacher, the linear evaluation result of EfficientNet-B0 is very close to the teacher, 
while the number of parameters of EfficientNet-B0 is only 9.4\%/16.3\% of ResNet-101/ResNet-50.

\begin{table*}[ht]
    \centering
    \caption{Object detection and instance segmentation results on VOC07 test and COCO val2017 with ResNet-34 as backbone. \ddag  means our implementation. Subscript in green represents the improvement compared to MoCo-V2 baseline.}
    \setlength{\tabcolsep}{1.5mm}{
    \begin{tabular}{c|c|c|ccc|ccc|ccc}
        \whline
        \multirow{3}{*}{\textbf{S}} & \multirow{3}{*}{\textbf{T}} & \multirow{3}{*}{\textbf{Method}} & \multicolumn{6}{c|}{\textbf{Object Detection}} &\multicolumn{3}{c}{\textbf{Instance Segmentation}}\\ 
        \cline{4-12}
        & & & \multicolumn{3}{c|}{\textbf{VOC}} & \multicolumn{3}{c|}{\textbf{COCO}} & \multicolumn{3}{c}{\textbf{COCO}}\\ 
        \cline{4-12}
        & & & $\bm{AP^{bb}}$ & $\bm{AP^{bb}_{50}}$ & $\bm{AP^{bb}_{75}}$ &  $\bm{AP^{bb}}$ & $\bm{AP^{bb}_{50}}$ & $\bm{AP^{bb}_{75}}$ &  $\bm{AP^{mk}}$ & $\bm{AP^{mk}_{50}}$ & $\bm{AP^{mk}_{75}}$\\
        \hline
        \multirow{7}{*}{\tabincell{c}{{}\\ \\ R-34}} & $\times$ & MoCo-V2$\ddag$ & 53.6 & 79.1 & 58.7 & 38.1 & 56.8 & 40.7 & 33.0 & 53.2 & 35.3\\
        \cdashline{2-12}
        & \multirow{2}{*}{\tabincell{c}{\\ R-50}}& SEED \cite{fang2021seed} & 53.7 & 79.4 & 59.2 & 38.4 & 57.0 & 41.0 & 33.3 & 53.2 & 35.3\\
        & &\tabincell{c}{\textbf{DisCo (ours)}\\ {}}& \tabincell{c}{\textbf{56.5}\\ \greenp{2.9$\uparrow$}} & \tabincell{c}{\textbf{80.6}\\ \greenp{1.5$\uparrow$}} & \tabincell{c}{\textbf{62.5}\\ \greenp{3.8$\uparrow$}} & \tabincell{c}{\textbf{40.0}\\ \greenp{1.9$\uparrow$}} & \tabincell{c}{\textbf{59.1}\\ \greenp{2.3$\uparrow$}} & \tabincell{c}{\textbf{43.4}\\ \greenp{2.7$\uparrow$}} & \tabincell{c}{\textbf{34.9}\\ \greenp{1.9$\uparrow$}} & \tabincell{c}{\textbf{56.3}\\ \greenp{3.1$\uparrow$}} & \tabincell{c}{\textbf{37.1}\\ \greenp{1.8$\uparrow$}}\\
        \cdashline{2-12}
        & \multirow{2}{*}{\tabincell{c}{\\ R-101}} & SEED \cite{fang2021seed} & 54.1 & 79.8 & 59.1 & 38.5 & 57.3 & 41.4 & 33.6 & 54.1 & 35.6 \\
        & &\tabincell{c}{\textbf{DisCo (ours)}\\ {}}& \tabincell{c}{\textbf{56.1}\\ \greenp{2.5$\uparrow$}} & \tabincell{c}{\textbf{80.3}\\ \greenp{1.2$\uparrow$}} & \tabincell{c}{\textbf{61.8}\\ \greenp{3.1$\uparrow$}} & \tabincell{c}{\textbf{40.0}\\ \greenp{1.9$\uparrow$}} & \tabincell{c}{\textbf{59.1}\\ \greenp{2.3$\uparrow$}} & \tabincell{c}{\textbf{43.2}\\ \greenp{2.5$\uparrow$}} & \tabincell{c}{\textbf{34.7}\\ \greenp{1.9$\uparrow$}} & \tabincell{c}{\textbf{55.9}\\ \greenp{2.7$\uparrow$}} & \tabincell{c}{\textbf{37.4}\\ \greenp{1.8$\uparrow$}}\\
        \cdashline{2-12}
        & \multirow{2}{*}{\tabincell{c}{\\ R-152}} & SEED \cite{fang2021seed} & 54.4 & 80.1 & 59.9 & 38.4 & 57.0 & 41.0 & 33.3 & 53.7 & 35.3\\
        & &\tabincell{c}{\textbf{DisCo (ours)}\\ {}} & \tabincell{c}{\textbf{56.6}\\ \greenp{3.0$\uparrow$}} & \tabincell{c}{\textbf{80.8} \\ \greenp{1.7$\uparrow$}} & \tabincell{c}{\textbf{63.4}\\ \greenp{5.7$\uparrow$}} & \tabincell{c}{\textbf{39.4}\\ \greenp{1.3$\uparrow$}} & \tabincell{c}{\textbf{58.7}\\ \greenp{1.9$\uparrow$}} & \tabincell{c}{\textbf{42.7}\\ \greenp{2.0$\uparrow$}} & \tabincell{c}{\textbf{34.4}\\ \greenp{1.4$\uparrow$}} & \tabincell{c}{\textbf{55.4}\\ \greenp{2.2$\uparrow$}} & \tabincell{c}{\textbf{36.7}\\ \greenp{1.4$\uparrow$}}\\
        \whline
    \end{tabular}
    }
    \label{tab:transfer-detection-segmentation}
\end{table*}

\subsection{Semi-supervised Linear Evaluation} 
\begin{figure}[ht]
    \centering
    \vspace{-9pt}
    \includegraphics[width=0.92\linewidth]{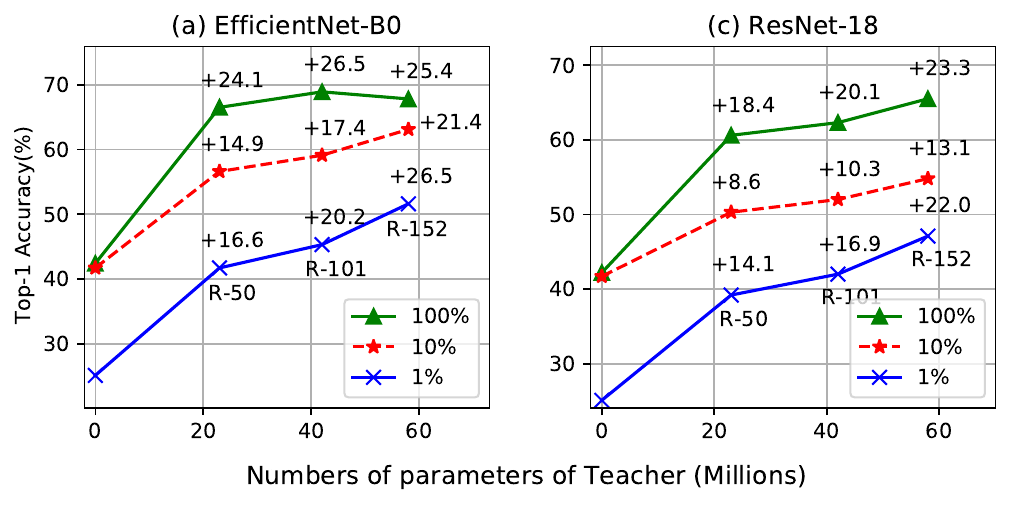}
    \vspace{-8pt}
    \caption{ImageNet top-1 accuracy (\%) of semi-supervised linear evaluation with 1\%, 10\% and 100\% training data. Points where the number of teacher network parameters are 0 are the results of the MoCo-V2 without distillation.}
    \vspace{-18pt}
    \label{fig:limited-labels}
\end{figure}

Following SEED, we evaluate our method under the semi-supervised setting. 
Two 1\% and 10\% sampled subsets of ImageNet training data ($\sim$12.8 and $\sim$128 images per class respectively) \cite{chen2020simple} are used for fine-tuning the student models. 
As is shown in Figure \ref{fig:limited-labels}, student models distilled by DisCo outperform baseline under any amount of labeled data. Furthermore, DisCo also shows the consistency under different fractions of annotations, that is, students always benefit from larger models as teachers. More labels will be helpful to improve the final performance of the student model, which is expected.

\begin{figure}[ht]
    \centering
    \includegraphics[width=0.88\linewidth]{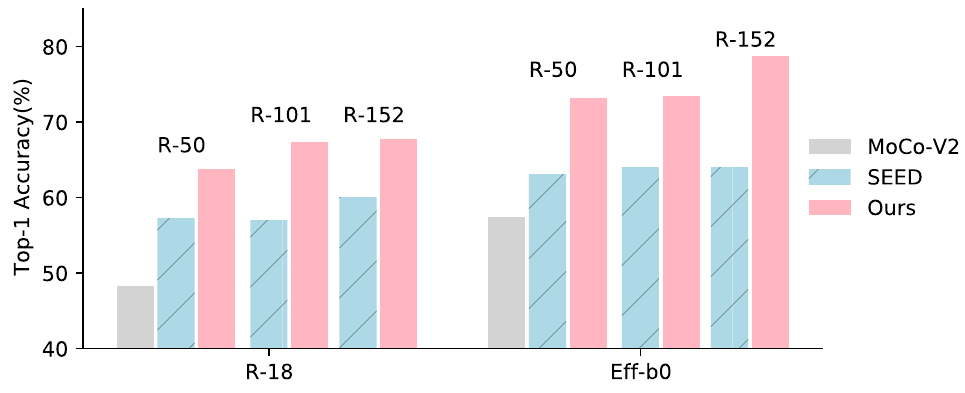}
    \caption{Top-1 accuracy of students transferred to Cifar100 without and with distillation from different teachers.}
    \label{fig:transfer-cifar100}
\end{figure}
\subsection{Transfer to Cifar10/Cifar100}
In order to analyze the generalization of representations obtained by DisCo, we further conduct linear evaluation on Cifar10 and Cifar100 with ResNet-18/EfficientNet-B0 as student and ResNet-50/ResNet101/ResNet152 as a teacher. Since the image resolution of the Cifar dataset is $32 \times 32$, all the images are resized to $224\times 224$ with bicubic re-sampling before feeding into the model, following \cite{fang2021seed}. The results are shown in Figure \ref{fig:transfer-cifar100}, it can be seen that the proposed DisCo surpasses the MoCo-V2 baseline by a large margin with different student and teacher architectures on and Cifar100. In addition, our method also has a significant improvement compared the-state-of-art method SEED. It is worth noting that as the teacher becomes better, the improvement brought by DisCo is more obvious. The performance trend on Cifar10 is consistent with that on Cifar100, see section 2 in the supplementary material for details.


\subsection{Transfer to Detection and Segmentation}

We also conduct experiments on detection and segmentation tasks for generalization analysis. C4 based Faster R-CNN \cite{ren2015faster} are used for objection detection on VOC and Mask R-CNN \cite{he2017mask} are used for objection detection and instance segmentation on COCO. The results 
are shown in Table ~\ref{tab:transfer-detection-segmentation}. On object detection, our method can bring obvious improvement on both VOC and COCO datasets. Furthermore, as SEED \cite{fang2021seed} claimed, the improvement on COCO is relatively minor compared to VOC since COCO training dataset has ~118k images while VOC has only ~16.5k training images, thus, the gain brought by weight initialization is relatively small. On the instance segmentation task, DisCo also shows superiority.

\subsection{Distilling BottleNeck Phenomenon}

In the self-supervised distillation stage, we first tried to distill small models with default MLP configuration of MoCo-V2 using ResNet-50 as a teacher, and the results are shown in Table \ref{tab:DisCo-SEED}, denoted by DisCo$^{*}$. It is worth noting that the dimensions of the hidden layer in DisCo$^{*}$ are exactly as same as SEED. It can be seen that compared to SEED, DisCo$^{*}$ shows superior results on EfficientNet-B0, and MobileNet-v3-Large, and has comparable results on ResNet-18. Then we expand the dimension of the hidden layer in the MLP of the student to be consistent with that of the teacher, that is, $2048D$, it can be seen that the results can be further improved, which is recorded in the third row. In particular, this expansion operation brings 3.5\% and 3.6\% gains for ResNet-18 and ResNet-34 respectively.
\begin{table}[ht]
    \centering
    \caption{Linear evaluation top-1 accuracy (\%) on ImageNet.}
    \setlength{\tabcolsep}{0.96mm}{ 
    \begin{tabular}{c|cccccc}
        \whline
         \textbf{Method} &  \textbf{Eff-b0} & \textbf{Mob-v3} & \textbf{R-18} & \textbf{R-34}\\
         \hline
         SEED  & 61.3 & 55.2 & 57.6 & 58.5 \\
         DisCo* & 65.6 & 63.8 & 57.1 & 58.9\\
         \cdashline{1-5}
         DisCo & \textbf{66.5}\greenp{\small{0.9$\uparrow$}} &  \textbf{64.4}\greenp{\small{0.6$\uparrow$}} & \textbf{60.6}\greenp{\small{3.5$\uparrow$}} & \textbf{62.5}\greenp{\small{3.6$\uparrow$}}\\
         \whline
    \end{tabular}}
    \label{tab:DisCo-SEED}
\end{table}

\begin{figure}[htb]
    \centering
    \vspace{-10pt}
        \includegraphics[width=0.88\linewidth]{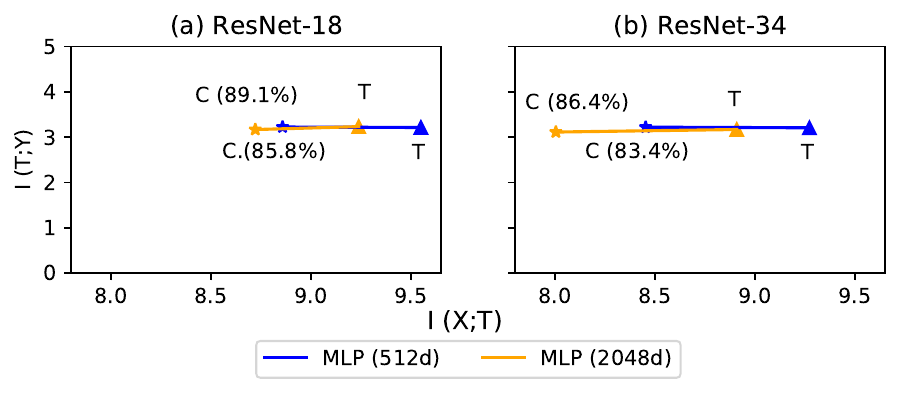}
    \vspace{-7pt}
    \caption{Mutual information paths from transition points to convergence points in the compression phase of training. $\textnormal{T}$ denotes transition points, and $\textnormal{C(X\%)}$ denotes convergent points with X\% top-1 accuracy on Cifar10. Points with similar $\textnormal{I(T;Y)}$ but smaller $\textnormal{I(X;T)}$ are better generalized.}
    \vspace{-11pt}
    \label{fig:mlp-ib}
    \vspace{-5pt}
\end{figure}

\noindent\textbf{Theoretical Analysis from IB perspective.} In Figure \ref{fig:mlp-ib}, on the downstream Cifar10 classification task, we visualize the \textit{compression phase} of ResNet-18/34 with different hidden dimensions distilled by the same teacher in the information plane. Following \cite{cheng2018evaluating}, we use binning strategy \cite{murphy2012machine} to estimate mutual information. It can be seen that when we adjust the hidden dimension in the MLP of ResNet-18 and ResNet-34 from $512D$ to $2048D$, the value of $I(X;T)$ becomes smaller while $I(T;Y)$ is basically unchanged, which suggests that enlarging the hidden dimension can make the student model more generalized in the setting of self-supervised transfer learning.

\subsection{Ablation Study}

In this section, we testify the effectiveness of two important modules in DisCo, i.e. the distillation loss and the expansion of the hidden dimension of MLP, and the results are shown in Table \ref{tab:AblationStudy}. It can be seen that distillation loss can bring about essential changes, and the result will be greatly improved. Even with only distillation loss, good results can be achieved. Furthermore, as the hidden dimension increases, the top-1 accuracy also increases, but when the dimension is already large, the growth trend will slow down. 

\begin{table}[ht]
    \centering
    \caption{Linear evaluation top-1 accuracy (\%) on ImageNet. MLP-d means the hidden dimension of MLP and - denotes the hidden layer of the MLP is directly removed.}
    \footnotesize
    \setlength{\tabcolsep}{0.24mm}{
    \begin{tabular}{ccc|ccc}
    \whline
    	\multicolumn{2}{c}{\textbf{Loss}} &
    	\multirow{2}{*}{\textbf{MLP-d}} &
    	\multirow{2}{*}{\textbf{Eff-b0}} &
    	\multirow{2}{*}{\textbf{Mob-v3}} & \multirow{2}{*}{\textbf{R-18}} \\
    	 $L_{co}$ & $L_{dis}$ & & & & \\
    \hline
        \textit{\textbf{Baseline}} & & & &\\
         \checkmark &  & 1280/1280/512 & 46.8 & 36.2 & 52.2 \\
         \cdashline{1-6}
         \multicolumn{2}{l}{\textit{\textbf{Effectiveness of loss}}} & & & & \\
         & \checkmark & 1280/1280/512 & 65.6 & 58.9 & 54.5 \\
         \checkmark & \checkmark & 1280/1280/512 & 65.6 & 63.7 & 57.1\\
         \cdashline{1-6}
         \multicolumn{2}{l}{\textit{\textbf{Effectiveness of MLP-d}}} & & & &\\
         \checkmark & \checkmark & -/-/- & 52.5 & 60.3 & 52.5\\
         \checkmark & \checkmark & 512/512/512 & 62.5 & 62.8 & 57.1\\
         \checkmark & \checkmark & 1024/1024/1024 & 65.0 & 63.8 & 59.2\\
         \checkmark & \checkmark & 2048/2048/2048 & \textbf{66.5} & \textbf{64.4} & \textbf{60.6}\\
        \whline
    \end{tabular}}
    \label{tab:AblationStudy}
\end{table}


\subsection{Comparison against other Distillation}

In order to verify the effectiveness of the proposed method, we compare with three widely used distillation schemes, namely, 1) \textit{Attention transfer} denoted by AT \cite{zagoruyko2016paying}, 2) \textit{Relational knowledge distillation} denoted by RKD \cite{park2019relational} 3) \textit{Knowledge distillation} denoted by KD \cite{hinton2015distilling}. AT and RKD are feature-based and relation-based respectively, which can be utilized during the self-supervised pre-training stage. KD is a logits-based method, which can only be used at the supervised fine-tuning stage. The comparison results are shown in Table \ref{tab:other-distillation}. \textit{Singe-Knowledge} means using one of these approaches individually, and it can be seen that all distillation approaches can bring improvement to the baseline but the gain from DisCo is the most significant, which indicates the knowledge that DisCo chosen to transfer and the way of transmission is indeed more effective. Then, we also try to transfer multi-knowledge from teacher to student by combining DisCo with other schemes. It can be seen that integrating DisCo with AT/RKD/KD can boost the performance a lot, which further proves the effectiveness of DisCo.

\begin{table}[htb]
    \centering
    \caption{Linear evaluation top-1 accuracy (\%) on ImageNet compared with different distillation methods.}
    \footnotesize
    \begin{tabular}{ccccc}
    \whline
    	 \textbf{Method} & \textbf{Eff-b0} & \textbf{Eff-b1} &\textbf{ Mob-v3 }& \textbf{R-18}\\
    \hline
    	 \textit{\textbf{Baseline}} & & & &\\
    	 MoCo-V2 & 46.8 & 48.4 & 36.2 & 52.2 \\
    	 \cdashline{1-5}
    	 \textit{\textbf{Single-Knowledge}} & & & &\\
    	 AT & 57.1 & 58.2 & 51.0 & 56.2\\
         RKD & 48.3 & 50.3 & 36.9 & 56.4\\
         KD & 46.5 & 48.5 & 37.3 & 51.5\\
         \textbf{DisCo (ours)} & \textbf{66.5} & \textbf{66.6} & \textbf{64.4} & \textbf{60.6}\\
         \cdashline{1-5}
         \textit{\textbf{Multi-Knowledge}} & & & &\\
         AT + DisCo & 66.7 & 66.3 & 64.1 & 60.0\\
         RKD + DisCo & \textbf{66.8} & \textbf{66.5} & 64.4 & \textbf{60.6}\\
         KD + DisCo & 65.8 & 65.9 & \textbf{65.2} & \textbf{60.6}\\
        \whline
    \end{tabular}
    \label{tab:other-distillation}
\end{table}

\subsection{More SSL Methods}

In order to demonstrate the versatility of our method, we further experiment with two SSL methods that are quite different from the MoCo-V2 baseline we used in the previous sections. i) SwAV is used to testify the compatibility towards the learning paradigm, in which the difference is measured between clusters instead of instances (see supplementary section 3); ii) DINO is used to testify the compatibility towards the backbone type, in which the encoder is a vision transformer instead of the commonly used CNN, as is shown in Table \ref{tab:vittabs}. It can be seen that DisCo is not limited to specific SSL methods, and can bring significant improvement under most of the popular SSL frameworks.



\begin{table}[ht]
    \centering
    \caption{Linear evaluation top-1 accuracy (\%) on ImageNet with DINO as testbed. ViT-small\cite{dosovitskiy2020image} and XCiT-small\cite{el2021xcit} are pre-trained by DINO for 100 epochs.}
    \small
    \setlength{\tabcolsep}{1.2mm}{ 
    \begin{tabular}{cc|ccccc}
    	\whline
    	\hline
    	 \textbf{Teacher Model} & \textbf{Acc} & \textbf{ViT-tiny} & \textbf{XCiT-tiny}  \\
    \hline
         - & - & 63.2 & 67.0  \\
         \cdashline{1-7}
    	 ViT-small & 77 & 68.4\greenp{5.2$\uparrow$} & - \\
         XCiT-small  & 77.8 & - & 71.1\greenp{4.1$\uparrow$}  \\
         \whline
    \end{tabular}}
    \label{tab:vittabs}
\end{table}

\section{Conclusion}

In this paper, we propose Distilled Contrastive Learning (DisCo) to remedy self-supervised learning on lightweight models. The proposed method constraints the final embedding of the lightweight student to be consistent with that of the teacher to maximally transmit the teacher's knowledge. DisCo is not limited to specific contrastive learning methods and can remedy student performance by a large margin.

{\small
\bibliographystyle{ieee_fullname}
\bibliography{egbib}

\begin{thebibliography}{10}\itemsep=-1pt

\bibitem{koohpayegani2020compress}
Soroush Abbasi~Koohpayegani, Ajinkya Tejankar, and Hamed Pirsiavash.
\newblock Compress: Self-supervised learning by compressing representations.
\newblock In {\em NeurIPS}, pages 12980--12992, 2020.

\bibitem{asano2020self}
Yuki~Markus Asano, Christian Rupprecht, and Andrea Vedaldi.
\newblock Self-labelling via simultaneous clustering and representation
  learning.
\newblock In {\em ICLR}, 2020.

\bibitem{caron2018deep}
Mathilde Caron, Piotr Bojanowski, Armand Joulin, and Matthijs Douze.
\newblock Deep clustering for unsupervised learning of visual features.
\newblock In {\em ECCV}, pages 132--149, 2018.

\bibitem{caron2019unsupervised}
Mathilde Caron, Piotr Bojanowski, Julien Mairal, and Armand Joulin.
\newblock Unsupervised pre-training of image features on non-curated data.
\newblock In {\em ICCV}, pages 2959--2968, 2019.

\bibitem{caron2020unsupervised}
Mathilde Caron, Ishan Misra, Julien Mairal, Priya Goyal, Piotr Bojanowski, and
  Armand Joulin.
\newblock Unsupervised learning of visual features by contrasting cluster
  assignments.
\newblock In {\em NeurIPS}, pages 9912--9924, 2020.

\bibitem{caron2021emerging}
Mathilde Caron, Hugo Touvron, Ishan Misra, Herv\'e J\'egou, Julien Mairal,
  Piotr Bojanowski, and Armand Joulin.
\newblock Emerging properties in self-supervised vision transformers.
\newblock 2021.

\bibitem{chen2020cross}
Defang Chen, Jian-Ping Mei, Yuan Zhang, Can Wang, Zhe Wang, Yan Feng, and Chun
  Chen.
\newblock Cross-layer distillation with semantic calibration.
\newblock 2020.

\bibitem{chen2020simple}
Ting Chen, Simon Kornblith, Mohammad Norouzi, and Geoffrey Hinton.
\newblock A simple framework for contrastive learning of visual
  representations.
\newblock In {\em ICML}, pages 1597--1607, 2020.

\bibitem{chen2020big}
Ting Chen, Simon Kornblith, Kevin Swersky, Mohammad Norouzi, and Geoffrey
  Hinton.
\newblock Big self-supervised models are strong semi-supervised learners.
\newblock In {\em NeurIPS}, pages 22243--22255, 2020.

\bibitem{chen2020mocov2}
Xinlei Chen, Haoqi Fan, Ross Girshick, and Kaiming He.
\newblock Improved baselines with momentum contrastive learning.
\newblock In {\em CVPR}, pages 9729--9738, 2020.

\bibitem{chen2020exploring}
Xinlei Chen and Kaiming He.
\newblock Exploring simple siamese representation learning.
\newblock In {\em arXiv preprint arXiv:2011.10566}, 2020.

\bibitem{cheng2018evaluating}
Hao Cheng, Dongze Lian, Shenghua Gao, and Yanlin Geng.
\newblock Evaluating capability of deep neural networks for image
  classification via information plane.
\newblock In {\em ECCV}, pages 168--182, 2018.

\bibitem{doersch2015unsupervised}
Carl Doersch, Abhinav Gupta, and Alexei~A Efros.
\newblock Unsupervised visual representation learning by context prediction.
\newblock In {\em ICCV}, pages 1422--1430, 2015.

\bibitem{dosovitskiy2020image}
Alexey Dosovitskiy, Lucas Beyer, Alexander Kolesnikov, Dirk Weissenborn,
  Xiaohua Zhai, Thomas Unterthiner, Mostafa Dehghani, Matthias Minderer, Georg
  Heigold, Sylvain Gelly, et~al.
\newblock An image is worth 16x16 words: Transformers for image recognition at
  scale.
\newblock {\em arXiv preprint arXiv:2010.11929}, 2020.

\bibitem{dosovitskiy2015discriminative}
Alexey Dosovitskiy, Philipp Fischer, Jost~Tobias Springenberg, Martin
  Riedmiller, and Thomas Brox.
\newblock Discriminative unsupervised feature learning with exemplar
  convolutional neural networks.
\newblock volume~38, pages 1734--1747, 2015.

\bibitem{el2021xcit}
Alaaeldin El-Nouby, Hugo Touvron, Mathilde Caron, Piotr Bojanowski, Matthijs
  Douze, Armand Joulin, Ivan Laptev, Natalia Neverova, Gabriel Synnaeve, Jakob
  Verbeek, et~al.
\newblock Xcit: Cross-covariance image transformers.
\newblock 2021.

\bibitem{everingham2010pascal}
Mark Everingham, Luc Van~Gool, Christopher~KI Williams, John Winn, and Andrew
  Zisserman.
\newblock The pascal visual object classes challenge.
\newblock volume~88, pages 303--338, 2010.

\bibitem{fang2021seed}
Zhiyuan Fang, Jianfeng Wang, Lijuan Wang, Lei Zhang, Yezhou Yang, and Zicheng
  Liu.
\newblock Seed: Self-supervised distillation for visual representation.
\newblock In {\em ICLR}, 2021.

\bibitem{grill2020bootstrap}
Jean-Bastien Grill, Florian Strub, Florent Altch\'{e}, Corentin Tallec, Pierre
  Richemond, Elena Buchatskaya, Carl Doersch, Bernardo Avila~Pires, Zhaohan
  Guo, Mohammad Gheshlaghi~Azar, Bilal Piot, koray kavukcuoglu, Remi Munos, and
  Michal Valko.
\newblock Bootstrap your own latent: A new approach to self-supervised
  learning.
\newblock In {\em NeurIPS}, pages 21271--21284, 2020.

\bibitem{he2019moco}
Kaiming He, Haoqi Fan, Yuxin Wu, Saining Xie, and Ross Girshick.
\newblock Momentum contrast for unsupervised visual representation learning.
\newblock In {\em CVPR}, pages 9729--9738, 2020.

\bibitem{he2017mask}
Kaiming He, Georgia Gkioxari, Piotr Doll{\'a}r, and Ross Girshick.
\newblock Mask r-cnn.
\newblock In {\em ICCV}, pages 2961--2969, 2017.

\bibitem{he2016deep}
Kaiming He, Xiangyu Zhang, Shaoqing Ren, and Jian Sun.
\newblock Deep residual learning for image recognition.
\newblock In {\em CVPR}, pages 770--778, 2016.

\bibitem{henaff2020data}
Olivier Henaff.
\newblock Data-efficient image recognition with contrastive predictive coding.
\newblock In {\em ICML}, pages 4182--4192, 2020.

\bibitem{hinton2015distilling}
Geoffrey Hinton, Oriol Vinyals, and Jeff Dean.
\newblock Distilling the knowledge in a neural network.
\newblock In {\em NeurIPSW}, 2015.

\bibitem{howard2019searching}
Andrew Howard, Mark Sandler, Grace Chu, Liang-Chieh Chen, Bo Chen, Mingxing
  Tan, Weijun Wang, Yukun Zhu, Ruoming Pang, Vijay Vasudevan, et~al.
\newblock Searching for mobilenetv3.
\newblock In {\em ICCV}, pages 1314--1324, 2019.

\bibitem{komodakis2018unsupervised}
Nikos Komodakis and Spyros Gidaris.
\newblock Unsupervised representation learning by predicting image rotations.
\newblock In {\em ICLR}, 2018.

\bibitem{krizhevsky2009learning}
Alex Krizhevsky and Geoffrey Hinton.
\newblock Learning multiple layers of features from tiny images.
\newblock Citeseer, 2009.

\bibitem{lin2014microsoft}
Tsung-Yi Lin, Michael Maire, Serge Belongie, James Hays, Pietro Perona, Deva
  Ramanan, Piotr Doll{\'a}r, and C~Lawrence Zitnick.
\newblock Microsoft coco: Common objects in context.
\newblock In {\em ECCV}, pages 740--755, 2014.

\bibitem{liu2019knowledge}
Yufan Liu, Jiajiong Cao, Bing Li, Chunfeng Yuan, Weiming Hu, Yangxi Li, and
  Yunqiang Duan.
\newblock Knowledge distillation via instance relationship graph.
\newblock In {\em CVPR}, pages 7096--7104, 2019.

\bibitem{murphy2012machine}
Kevin~P Murphy.
\newblock {\em Machine learning: a probabilistic perspective}.
\newblock MIT press, 2012.

\bibitem{noroozi2016unsupervised}
Mehdi Noroozi and Paolo Favaro.
\newblock Unsupervised learning of vis,ual representations by solving jigsaw
  puzzles.
\newblock In {\em ECCV}, pages 69--84, 2016.

\bibitem{park2019relational}
Wonpyo Park, Dongju Kim, Yan Lu, and Minsu Cho.
\newblock Relational knowledge distillation.
\newblock In {\em CVPR}, pages 3967--3976, 2019.

\bibitem{pathak2016context}
Deepak Pathak, Philipp Krahenbuhl, Jeff Donahue, Trevor Darrell, and Alexei~A
  Efros.
\newblock Context encoders: Feature learning by inpainting.
\newblock In {\em CVPR}, pages 2536--2544, 2016.

\bibitem{ren2015faster}
Shaoqing Ren, Kaiming He, Ross Girshick, and Jian Sun.
\newblock Faster r-cnn: Towards real-time object detection with region proposal
  networks.
\newblock volume~39, pages 1137--1149, 2015.

\bibitem{romero2014fitnets}
Adriana Romero, Nicolas Ballas, Samira~Ebrahimi Kahou, Antoine Chassang, Carlo
  Gatta, and Yoshua Bengio.
\newblock Fitnets: Hints for thin deep nets.
\newblock In {\em ICLR}, 2014.

\bibitem{ILSVRC15}
Olga Russakovsky, Jia Deng, Hao Su, Jonathan Krause, Sanjeev Satheesh, Sean Ma,
  Zhiheng Huang, Andrej Karpathy, Aditya Khosla, Michael Bernstein,
  Alexander~C. Berg, and Li Fei-Fei.
\newblock {ImageNet Large Scale Visual Recognition Challenge}.
\newblock volume 115, pages 211--252, 2015.

\bibitem{shwartz2017opening}
Ravid Shwartz-Ziv and Naftali Tishby.
\newblock Opening the black box of deep neural networks via information.
\newblock 2017.

\bibitem{tan2019efficientnet}
Mingxing Tan and Quoc Le.
\newblock Efficientnet: Rethinking model scaling for convolutional neural
  networks.
\newblock In {\em ICML}, pages 6105--6114, 2019.

\bibitem{tian2019contrastive}
Yonglong Tian, Dilip Krishnan, and Phillip Isola.
\newblock Contrastive representation distillation.
\newblock In {\em ICLR}, 2020.

\bibitem{tishby2000information}
Naftali Tishby, Fernando~C Pereira, and William Bialek.
\newblock The information bottleneck method.
\newblock 2000.

\bibitem{wang2020enhancing}
Jinpeng Wang, Yuting Gao, Ke Li, Xinyang Jiang, Xiaowei Guo, Rongrong Ji, and
  Xing Sun.
\newblock Enhancing unsupervised video representation learning by decoupling
  the scene and the motion.
\newblock 2020.

\bibitem{wang2020removing}
Jinpeng Wang, Yuting Gao, Ke Li, Yiqi Lin, Andy~J Ma, and Xing Sun.
\newblock Removing the background by adding the background: Towards background
  robust self-supervised video representation learning.
\newblock 2020.

\bibitem{xu2020knowledge}
Guodong Xu, Ziwei Liu, Xiaoxiao Li, and Chen~Change Loy.
\newblock Knowledge distillation meets self-supervision.
\newblock In {\em ECCV}, pages 588--604, 2020.

\bibitem{zagoruyko2016paying}
Sergey Zagoruyko and Nikos Komodakis.
\newblock Paying more attention to attention: Improving the performance of
  convolutional neural networks via attention transfer.
\newblock In {\em ICLR}, 2017.

\bibitem{zbontar2021barlow}
Jure Zbontar, Li Jing, Ishan Misra, Yann LeCun, and St{\'e}phane Deny.
\newblock Barlow twins: Self-supervised learning via redundancy reduction.
\newblock 2021.

\end{thebibliography}
}

\end{document}